\documentclass[journal]{vgtc}                     

\onlineid{1774}

\usepackage{dblfloatfix}
\usepackage{amsmath}

\vgtccategory{Research}

\vgtcpapertype{application/design study}

\title{OW-CLIP: Data-Efficient Visual Supervision for Open-World Object Detection via Human-AI Collaboration}

\author{%
  Junwen Duan,
  Wei Xue,
  Ziyao Kang,
  \authororcid{Shixia Liu}{0000-0003-4499-6420}, and  
  \authororcid{Jiazhi Xia$^*$}{0000-0003-4629-6268}
}

\authorfooter{
  \item Junwen Duan, Wei Xue, Ziyao Kang, Jiazhi Xia are with the School of      Computer Science and Engineering, Central South University.
  	E-mail: \{jwduan,234711076,kangzy,xiajiazhi\}@csu.edu.cn. 
  \item Shixia Liu is with the School of Software, Tsinghua University.
  	E-mail: shixia@tsinghua.edu.cn.
  \item Jiazhi Xia is the corresponding author.
}

\abstract{Open-world object detection (OWOD) extends traditional object detection to identifying both known and unknown object, necessitating continuous model adaptation as new annotations emerge. Current approaches face significant limitations:
1) data-hungry training due to reliance on a large number of crowdsourced annotations, 
2) susceptibility to "partial feature overfitting," and 
3) limited flexibility due to required model architecture modifications. 
To tackle these issues, we present OW-CLIP, a visual analytics system that provides curated data and enables data-efficient OWOD model incremental training.
OW-CLIP implements plug-and-play multimodal prompt tuning tailored for OWOD settings and introduces a novel "Crop-Smoothing" technique to mitigate partial feature overfitting. 
To meet the data requirements for the training methodology, we propose dual-modal data refinement methods that leverage large language models and cross-modal similarity for data generation and filtering. Simultaneously, we develope a visualization interface that enables users to explore and deliver high-quality annotations—including class-specific visual feature phrases and fine-grained differentiated images.
Quantitative evaluation demonstrates that OW-CLIP achieves competitive performance at 89\% of state-of-the-art performance while requiring only 3.8\% self-generated data, while outperforming SOTA approach when trained with equivalent data volumes. 
A case study shows the effectiveness of the developed method and the improved annotation quality of our visualization system. 

}


\teaser{
  \centering 
  \includegraphics[width=\linewidth]{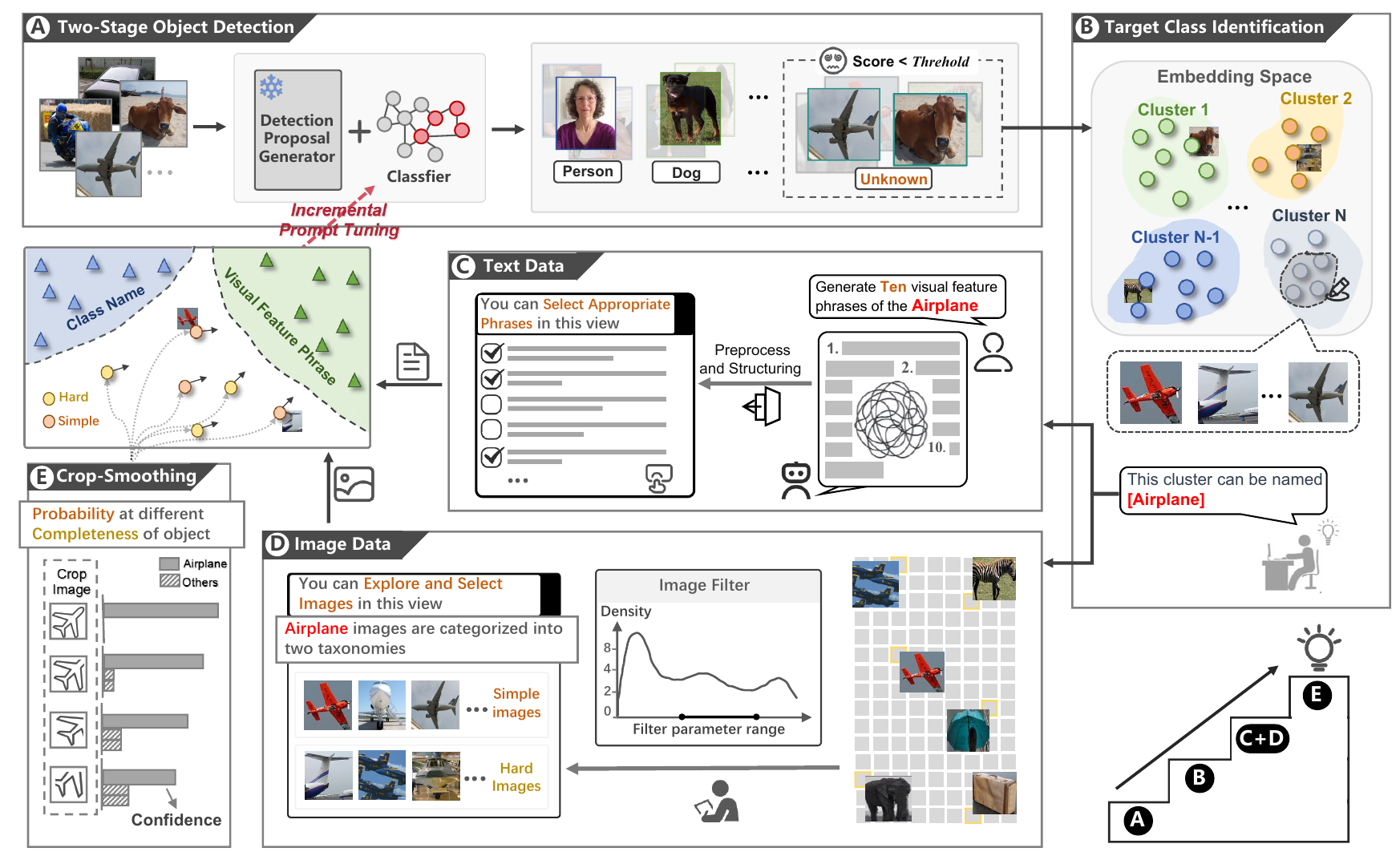}
  \caption{OW-CLIP system overview. (A) The system adopts two-stage
    object detection framework. (B) All unknown objects will be automatically clustered based on their image encoding results. Users determine potential unknown classes for annotation and inputs the corresponding names into the system. (C) Using a large language model to generate visual feature phrases, which are then transformed into a structured format and displayed on the visualization interface.  (D) All training images will undergo fine-grained differentiation by users. Moreover, we implement cross-modal similarity filtering to eliminate irrelevant images. (E) Based on fine-grained image annotation, a novel "Crop-Smoothing" technique is applied to incremental prompt tuning to mitigate the "partial feature overfitting".
  }
  \label{fig:overview}
}

\graphicspath{{figs/}{figures/}{pictures/}{images/}{./}} 

\usepackage{tabu}                      
\usepackage{booktabs}                  
\usepackage{lipsum}                    
\usepackage{mwe}                       

\usepackage{subcaption}
\usepackage{makecell}
\usepackage{multirow}

\begin{document}



\firstsection{Introduction}
\maketitle

\begin{figure}[tb]
  \centering 
  \includegraphics[width=0.95\columnwidth]{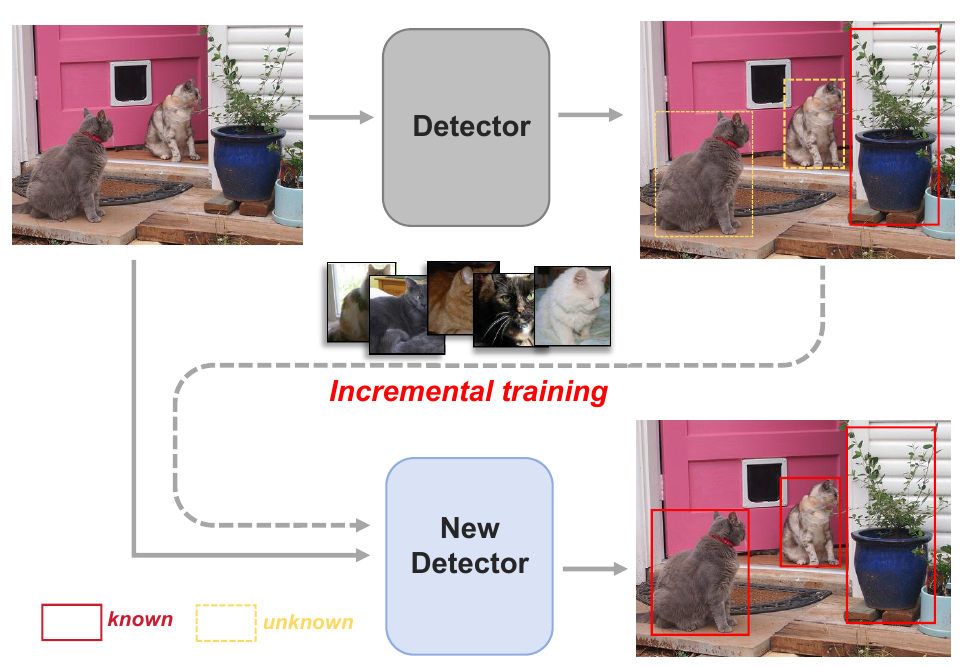}
  \caption{An example from the OWOD task: the model initially fails to recognize the object “cat” and classifies it as unknown. After receiving the corresponding annotation and retraining, the model is able to correctly identify it. }
  \label{fig:OWOD-example}
\end{figure}

Open-world object detection (OWOD)~\cite{joseph2021towards} extends traditional object detection by relaxing the closed-world assumption. In OWOD, as shown in Figure~\ref{fig:OWOD-example}, models not only detect known classes predefined in the training data but also identify potential unknown objects never encountered during training. These models need to continuously acquire annotations for newly discovered objects and incrementally update their knowledge. This paradigm shift addresses the limitations of conventional object detection systems, which struggle in dynamic, real-world environments where new object classes frequently emerge.

Recent year have witnessed substantial progress in OWOD research, including unknown object discovery~\cite{zheng2022towards,de2017guesswhat,fini2021unified} and incremental learning ~\cite{zhuang2024ds,wu2019large}.
However, there remain three critical limitations that hinder the widespread adoption and effectiveness of current OWOD systems:

\textit{Limited Data Quality}: Existing methods heavily rely on a large amount of annotations for training unknown classes, which is often costly and time-consuming in open-world contexts~\cite{shu2018unseen,bendale2015towards,joseph2021towards}. These annotations typically include only class name and low-quaily images, containing limited information. This paucity of descriptive and fine-grained data forces the model to rely on a large number of crowdsourced labels to achieve satisfactory performance, hindering efficient training and making rapid adaptation to new object classes impractical.

\begin{figure}[tb]
  \centering 
  \includegraphics[width=0.95\columnwidth]{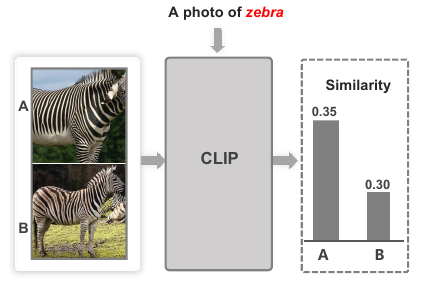}
  \caption{Visualization of partial feature overfitting. The classification confidence for an entire object is lower than that for its local regions. Image A (partial zebra) produces higher similarity scores than Image B (complete zebra), demonstrating CLIP's overreliance on local features. }
  \label{fig:zebra-compare}
\end{figure}

\textit{Partial Feature Overfitting}: Vision-language models like CLIP~\cite{radford2021learning} and ALIGN~\cite{li2021align} excel at learning generalized representations by training on large-scale image-text pairs in a contrastive manner, and have been adopted by diverse downstream tasks\cite{wang2022cris,rao2022denseclip,yao2022detclip,jamonnak2023ow,khattak2023maple}.
However, our observation reveals a limitation referred to as "partial feature overfitting", where these models overly rely on local salient features while neglecting global structure and contextual information, as illustrated in Figure~\ref{fig:zebra-compare}, leading to poor generalization in OWOD. 
No effective solutions have been proposed, as most existing works\cite{yao2022detclip,jamonnak2023ow,khattak2023maple}  focus on improving model architectures, neglecting the critical role of training data quality.

\textit{Limited Model Flexibility}: OWOD models need to continuously acquire annotations for newly discovered objects and incrementally update their knowledge. To achieve advanced performance in OWOD tasks, most existing approaches\cite{joseph2021towards,gupta2022ow,wang2023random} require significant modifications to the model architecture and repeatedly fine-tune using previously seen training data. These modifications can enhance performance for specific scenarios but limit the model's flexibility and adaptability to diverse and evolving open-world environments.

To address these challenges, we develop OW-CLIP, a visual analytics system that leverages human-AI collaboration to 1) provide curated data for new classes; 
2) enable data-efficient incremental training of CLIP models in OWOD.
Specifically, given the continuous knowledge-updating requirement of the OWOD task, OW-CLIP implements a plug-and-play prompt tuning for the CLIP model. This approach enables quick adaptation to unknown object classes without altering the architecture and previously trained parameters, ensuring both flexibility and scalability in OWOD tasks. At this process, we introduce a novel "Crop-Smoothing" technique for the first time, which dynamically adjusts the final predicted logits vector for the target class according to the completeness of the objects in the image, thereby  mitigating feature overfitting of the model. 
To address the critical data quality bottleneck in open-world object detection
, we develop a dual-modal annotation visualization as the core of OW-CLIP.
a) On the text modality: OW-CLIP incorporates a tailored workflow that leverages a large language model (LLM) as an "assistant" for automatically generating rich visual feature phrases for target unknown class.
The raw string outputs from the LLM are processed and transformed into a structured phrase list. These phrases provide detailed class-specific attributes beyond simple class name in prior works~\cite{joseph2021towards,gupta2022ow,zhao2023revisiting,wang2023random}.
This approach eliminates the need for manual description while providing descriptive text, thus greatly reducing the annotation burden on users.
b) On the image modality: We propose the cross-modal similarity guided image filtering method that filters out irrelevant image. All training images are differentiated as "Simple" and "Hard" (as detailed in Section~\ref{sec:Data-Refinement}), which ensures that the model learns not only the features of the object but also gains information about the completeness of the object during training. 
With the support of the visualization interface, users provide refined training data for potential unknown classes with minimal human effort, including class-specific \textit{Visual Feature Phrases} and \textit{Fine-grained Differentiated Images}.
These annotations are then used for model incremental training. We use class feature phrases as text vector initialization, and based on fine-grained differentiated images, dual-path Crop-Smoothing processing strategy is employed to handle both Simple and Hard image.

Quantitative evaluation experiment demonstrates that OW-CLIP achieves competitive performance at 89\% of state-of-the-art(SOTA) method while requiring only 3.8\% our curated data. When trained with equivalent data volumes, our framework outperforms current SOTA performance. A case study shows the effectiveness of the developed method and the improved annotation quality of our system. Additionally, user study further confirms the usability of our system.

In summary, the main contributions of this work are:
\begin{itemize}
    \item A data-efficient OWOD method that utilizes only a fraction of the annotation to achieve performance comparable to high-performing models trained on traditional OWOD datasets.
    \item A visualization system designed to provide high-quality text and image annotations with minimal human effort.
    \item A novel technique called "Crop-Smoothing" to mitigate partial feature overfitting issue.
    \item Interactive incremental prompt tuning to maintain the model's flexibility and scalability tailored for OWOD settings.
\end{itemize}

\section{RELATED WORK}

\subsection{Open-World Object Detection}

The open-world object detection problem was first proposed by Joseph and Khan~\cite{joseph2021towards}, and recent progress has been significant. Some studies have focused on improving data quality and utilization to enhance open-world learning. DetCLIP~\cite{yao2022detclip} creates a text dictionary to provide prior knowledge for each class, building relationships among concepts to facilitate learning. ABR~\cite{liu2023augmented} stores and replays complete object images to avoid foreground shift issues. Other research has explored model architectural enhancements to improve adaptability. Methods like L2P~\cite{wang2022learning} and ECLIPSE~\cite{kim2024eclipse} freeze base model and fine-tune prompt embeddings, addressing catastrophic forgetting and supporting incremental learning. The latest research, RandBox~\cite{wang2023random} uses debiasing through randomization, improving detection accuracy with random proposals and matching scores.
Furthermore, to alleviate hallucination in vision-language models, Liu et al.~\cite{liu2024paying} propose an effective approach that amplifies image token attention and debiases text-dominated outputs, significantly improving the visual grounding accuracy.
These techniques successfully detect both known and unknown classes.

The aforementioned methods rely on a large number of crowdsourced annotations for training unknown classes, which is often prohibitively costly and time-consuming in open-world scenarios. Compared with them, OW-CLIP achieves satisfactory method using only a fraction of curated annotations and outperforms current SOTA performance with equivalent data volume.

\subsection{Visualization for Annotation Quality
Improvement}  

Many existing datasets rely on non-expert annotations, leading to data quality issues. Improving annotation quality is a major focus in visualization research. 

Recently, VIS+AI~\cite{lu2024agentlens} presents a visionary framework for combining visualization and AI, offering insights into how human-AI collaboration can support complex data analysis tasks. LabelInspect~\cite{liu2018interactive} introduces an interactive framework for experts to validate ambiguous labels and identify unreliable annotators. Similarly, MutualDetector~\cite{chen2021towards} allows users to refine labels and detected objects iteratively, enhancing both label extraction and object detection through user feedback. However, datasets like ImageNet~\cite{russakovsky2015imagenet} lack annotator-related metadata, making it hard to address label noise. Xiang et al.~\cite{xiang2019interactive} tackled this with a trusted-item-based correction algorithm and an incremental t-SNE refinement method. These methods aim to improve data quality in existing datasets but aren't suited for incremental learning.

Beyond reducing noisy annotations, some studies refine basic annotations into high-quality data using external models and interactive human input. Jia et al.~\cite{jia2021towards} presents a visual explainable active learning framework to enhance zero-shot classification by guiding humans in defining attributes interactively. 
XNLI~\cite{feng2023xnli} is an insightful designed system that supports explainable visual data analysis by offering provenance-based query diagnosis, intelligent query revisions, and seamless human-AI collaboration.
However, these methods require extensive human involvement, limiting scalability.

Our work focuses on open-world object detection. Recently, OW-Adapter~\cite{jamonnak2023ow} was proposed for generating annotations for unknown classes in the OWOD problem, involving simple labeling and rough image selection, which led to suboptimal model performance. Unlike these methods, our work introduces a visualization analysis system that enhances data annotation and quality through human-computer interaction, providing rich visual features and refined images.

\section{DESIGN OF OW-CLIP}
This section presents the design of OW-CLIP. We begin with an in-depth requirement analysis that defines the key challenges and objectives. Next, the system overview provides a comprehensive look at the components and workflow. 
\subsection{Requirement Analysis}
\label{sec:System-Design}

The development of OW-CLIP was in collaboration with domain experts from diverse backgrounds.
The participants included Professor E1, researchers E2 and E3, who specialize in computer vision with a focus on OWOD, as well as Associate Professor E4 and researcher E5, whose expertise lies in artificial intelligence, particularly in NLP and multimodal domains.
To tackle the critical challenges in OWOD identified in our introduction, we conducted a series of semi-structured interviews with these experts, followed by brainstorming sessions to explore potential solutions. Through this process, we identified four key requirements for our visual analytics system, each directly addressing the limitations of existing OWOD methods.

\textbf{R1:}~\textit{Continuous Adaptation for OWOD with Minimal Data.}
Given the characteristics of the OWOD task, the model must continuously adapt to new classes using minimal additional data. E1 emphasized that existing methods rely on extensive crowdsourced annotations and full retraining, which hinders scalability. Therefore, a data-efficient continuous learning approach is essential for seamlessly integrating new data and updating object detectors in open-world scenarios.

\textbf{R2:}~\textit{Dual-Modal Annotation Support.} To tackle the challenge of data quality, our system must provide tools for generating and curating high-quality dual-modal training data for new object classes. This requirement is broken down into two sub-requirements:

\begin{enumerate}[label=\textbf{R2.\arabic*:},leftmargin=*]
  \item \textit{Fine-grained Image Differentiation.}
  Experts agreed that refining image quality plays a crucial role in improving model performance. E1 remarked, "In many image datasets, I've noticed that objects often appear incomplete, which can lead the model to learn the wrong things."
  In our current practice, our system must provide an intuitive interface for users to categorize images. 
 \item \textit{Visual Feature Phrase Exploration and Selection.}  
  Previous studies~\cite{jia2021towards,yao2022detclip,uehara2024learning} have shown that incorporating class feature descriptions or definitions in classification or detection tasks provides richer information beyond simple class names. Therefore, experts (E4 and E5) expressed that the system should provide such descriptive data to facilitate training.
\end{enumerate}

\textbf{R3:}~\textit{Annotation Efficiency Enhancement.} It is clearly impractical for users to manually extract the required training data from a vast number of unknown images and manually annotate detailed class feature descriptions. The visualization system needs to have the ability to organize, classify, filter images and automatically generated visual feature phrases for each object class.
\begin{enumerate}[label=\textbf{R3.\arabic*:},leftmargin=*]
  \item \textit{Unknown Object Cluster Visualization.}
  The system should offer a clear visual overview of unknown object clusters, allowing users to quickly identify and prioritize classes with distinct features or large sample sizes. E2 highlighted that given the overwhelming volume of images often present in datasets, this visualization must support interactive exploration and efficient selection of image clusters for annotation.
  \item \textit{Interactive Feature Phrase Refinement.}
  E4 commented: "When using an LLM as an 'assistant', it is important to consider that its output is a stream of text, which can make it inconvenient for users to explore."
  Users should be able to interactively refine and customize visual feature phrases through intuitive operations such as drag-and-drop or single-click selections. 
  \item \textit{Intelligent Image Recommendation.}
  To reduce the cognitive load of manual image sorting, experts agreed that the system should provide intelligent recommendations for images that are likely to belong to the same unknown class. This feature should support efficient organization and filtering of unknown examples throughout the annotation process.
\end{enumerate}

\textbf{R4:}~\textit{Comprehensive Model Performance Visualization.} 
The system should provide a clear, interactive visualization of the model's current detection capabilities. This overview should highlight both well-recognized classes and those requiring additional training, guiding users in their annotation efforts.

\subsection{System Overview}

Based on the these requirements, we developed a visualization system called OW-CLIP  to enable data-efficient incremental training of OWOD models (\textit{R1}).
As shown in Figure~\ref{fig:overview}, the entire system consists of two main modules: \textit{model incremental training} and \textit{visualization}. 

The model incremental training module adopts a two-stage object detection framework (Figure~\ref{fig:overview}A): the input image are first processed by a detection proposal generator to detect all potential objects, and the resulting proposals are then classified by a classifier. Proposals that cannot be matched to any known class are assigned the label "unknown". During the learning phase for these unknown classes, we implement a plug-and-play prompt tuning for the CLIP model, while introducing a novel "Crop-Smoothing" technology in this process.
The visualization module initially presents all proposals that the current model fails to recognize. The Labeling View enables users to identify potential unknown classes and assign appropriate names (\textit{R3.1}). Once the class label is defined, Dual-modal data refinement methods are employed to improve annotation efficiency (\textit{R3}), and users annotate textual data in the Feature Phrase Selection View and image data in the Image Selection View (\textit{R2}). The annotated textual data, along with images processed using the Crop-Smoothing technique, are utilized to incrementally update the CLIP. The model's performance will be displayed in the Results Display Panel (\textit{R4}).

\section{Model Incremental Training}
\label{section:Model-Reﬁnement}

The major goal of the developed OWOD method is to enable data-efficient incremental training utilizing a small number of curated annotations. Previous studies~\cite{wang2023detecting,liang2023open} have revealed a critical challenge in the evolution of the OWOD model: while detection proposal generation naturally extends to novel classes during incremental learning phases, the classification module exhibits persistent bias toward previously learned classes. 
Therefore, as shown in Figure~\ref{fig:overview}(A), we freeze the detection proposal generator while optimizing the classifier through multi-modal prompt tuning tailored for OWOD settings. 

\begin{figure}[t]
\centering
\includegraphics[width=0.95\linewidth]{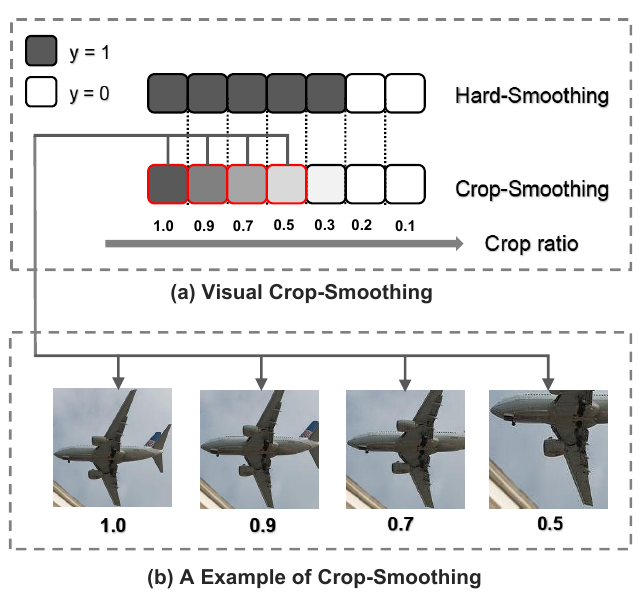}
\caption{Given an image of a complete object, we randomly crop it to generate several images. As the cropping ratio decreases, the object becomes increasingly incomplete, reducing the confidence for corresponding label while distributing higher probabilities to other labels during training.}
\label{CropSmoothing}
\end{figure}

\subsection{Crop-Smoothing}
\label{section:Crop-smoothing}

Crop-Smoothing is founded on the principle that classification confidence should positively correlate with the completeness of the object in the image. 
Figure~\ref{CropSmoothing}(a) illustrates our method, given a image, i.e., a detection proposal of a object, we randomly crop it to generate several images. The classification confidence assigned to the corresponding label is determined based on the cropping ratio. Assuming the cropping ratio is $\epsilon$, the ground truth is assigned a confidence of $D\epsilon$, and the remaining confidence $1-D\epsilon$ is equally distributed among the other labels. The confidence for each of these labels is  $\frac{1-D\epsilon}{Q-1}$ (where $\epsilon$ > ${\epsilon}_{min}$), and $Q$ denotes the total number of classes in the training process. Therefore, we define the loss function as shown in Equation~\ref{eq:Loss}.
\begin{equation}
       \label{eq:Loss} Loss=-\left[D\epsilon\cdot\log(\hat{y}_{j})+\sum_{i\neq {j}}\frac{1-D\epsilon}{Q-1}\cdot\log(\hat{y}_{i})\right]
\end{equation}
where $\hat{y}_{j}$ be the predicted probability for class $i$, $j$ be the ground truth label. 
Figure~\ref{CropSmoothing}(b) provides an example of applying Crop-Smoothing to an "airplane" Simple image.
This method dynamically adjusts the predicted logits for the target class based on the image's completeness. As a result, the model not only learns the intrinsic features of the object but also gains awareness of the completeness of the object in the image, improving robustness during training.

\begin{figure}[t]
  \centering 
  \includegraphics[width=0.95\columnwidth]{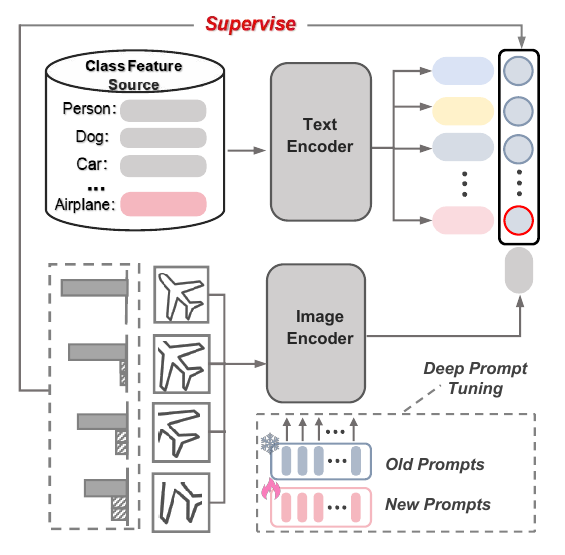}
  \caption{When new annotations for unknown classes become available, we perform multimodal prompt tuning on the image encoder and train the context vectors corresponding to the newly added classes. 
  }
  \label{fig:Tuning}
\end{figure}

\subsection{Multimodal Prompt Tuning for OWOD}
\label{section:prompt tuning}

Recent CLIP extensions have leveraged prompt tuning~\cite{zhou2022conditional,zhou2022learning,wasim2023vita,schick2020exploiting,lester2021power,li2021prefix} to enhance performance on downstream tasks. 
Unlike prior approaches, our method employs class feature phrases as text initialization. At the same time, we enhance training images with Crop-Smoothing. The learnable multimodal tokens generated in each training episode are incrementally added to the image encoder and the Class Feature Source without altering the previously parameter.

\textit{Update the CLIP image encoder.} 
We use deep prompt tuning mentioned in visual prompt tuning (VPT)~\cite{jia2022visual} to fine-tuning the CLIP image encoder. Specifically, we denote the input embeddings from the l-th multi-head attention model (MHA) module of the ViT-based image encoder in CLIP as $\{\mathbf{g}^{l},\mathbf{h}_{1}^{l},\mathbf{h}_{2}^{l},\cdots,\mathbf{h}_{N}^{l}\}$ , where $g^{l}$ denotes the [CLS] token embedding and $\mathbf{H}^{l}=\{\mathbf{h}_{1}^{l},\mathbf{h}_{2}^{l},\cdots,\mathbf{h}_{N}^{l}\}$ denotes the image patch embeddings. 
When new classes are introduced, we apply a freeze-and-tune strategy repeatedly. Deep prompt tuning appends learnable tokens $\mathbf{P}^{l} = \{\mathbf{p}_{1}^{l},\mathbf{p}_{2}^{l},\cdots,\mathbf{p}_{M}^{l}\}$ to the above token sequence in each ViT layer. Then
the l-th MHA module of t-th task process the input token as:
\begin{equation}
       \label{eq:CLIP_imageencoder} [\mathbf{g}^l, -,\mathbf{H}^l]=\mathrm{Layer}^l([\mathbf{g}^{l-1},\mathbf{P}_t^{l-1},\mathbf{H}^{l-1}])
\end{equation}
where we treat each set of prompts $\mathbf{P}_t^{l-1}$ as a
discrete task-specific module, solely devoted to the recognition of t-th task. The output embeddings of  $\{\mathbf{p}_1^l,\cdots,\mathbf{p}_M^l\}$  are discarded (denoted as -) and will not feed into the next layer.
Therefore, $\{\mathbf{p}_{1}^{l},\mathbf{p}_{2}^{l},\cdots,\mathbf{p}_{M}^{l}\}$ merely acts as a set of learnable parameters to adapt the MHA model.


\textit{Update the Class Feature Source.} 
The Class Feature Source stores contextual vectors corresponding to different classes. 
We maintain full parameter freezing of the CLIP text encoder to preserve its pre-trained linguistic knowledge while establishing learnable continuous vectors $\{\mathbf{c}_1, \dots, \mathbf{c}_K\} \in \mathbf{R}^d$ for multimodal representation learning.
Each class entry is initialized with phrase embeddings that retain lexical specificity through dimension-wise feature preservation.  This initialization strategy captures richer semantic and visual cues, leading to improved training effectiveness.


We set a classification confidence threshold t (a hyperparameter). Each detection proposal is matched against the learned feature representations of known classes, which are stored in the current class feature source. Proposals with matching scores above t are classified as known,
while those with scores below t are identified as unknown.
\begin{figure}[t]
\centering
\includegraphics[width=\linewidth]{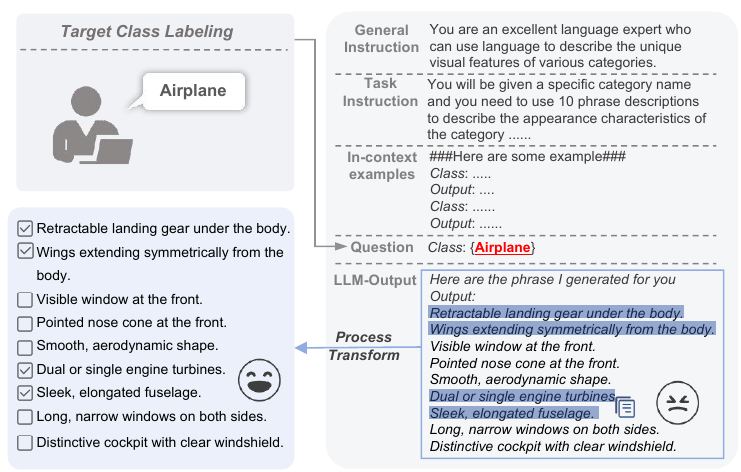}
\caption{The tailored workflow of OW-CLIP's LLM-powered phrase generation module. It leverages an LLM to generate visual feature phrases for target unknown classes, preprocesses the output into a structured list, and presents it in a visualization interface for efficient user selection.}
\label{fig:crop_LLMprocess}
\end{figure}

\section{OW-CLIP VISUALIZATION}
Traditional datasets lack the granular attribute descriptors and fine-grained images. To address the critical data quality bottleneck in open-world object detection,  we introduce a noval interactive visualization that 1) implements dual-modal data refinement methods, and 2) develops a visualization interface that enables users to explore and refine data. In the final subsection, we describe how these annotations drive our multi-modal prompt tuning methodology.

\subsection{Dual-Modal Data Refinement Methods}
\label{sec:Data-Refinement}
To accelerate the annotation process and facilitate the exploration of large amounts of texts and images, we propose LLM-Driven Feature Phrase Generation for the text modality and Cross-Modal Similarity Guided Image Filtering for the image modality.

\textit{LLM-Driven Feature Phrase Generation.}
Large Language Models (LLMs)~\cite{GPT-4o,yang2024qwen2,guo2025deepseek} have garnered substantial scholarly attention for their exceptional text comprehension capabilities, facilitating their deployment across diverse downstream applications~\cite{de2023chatgpt,birhane2023science,li2024llava,wang2023gpt,ma2023large}. Consequently, we leverage LLMs to generate high-quality textual data—specifically, visual feature phrases. 
Specifically, OW-CLIP implements a sophisticated preprocessing pipeline, illustrated in Figure~\ref{fig:crop_LLMprocess}. The user-provided class label is integrated into a carefully crafted prompt template designed to guide the LLM in generating several distinctive visual feature phrases. 
Once the prompt are constructed, we feed the prompt into the LLM. 
Following the predefined output format for the LLM, the raw character streams produced by the model undergo automated processing and transformation into a structured, hierarchical list of phrases. 
This streamlined process eliminates the necessity for manual description while simultaneously providing contextually relevant and semantically meaningful textual data.
\begin{figure}[tp]
\centering
\includegraphics[width=\linewidth]{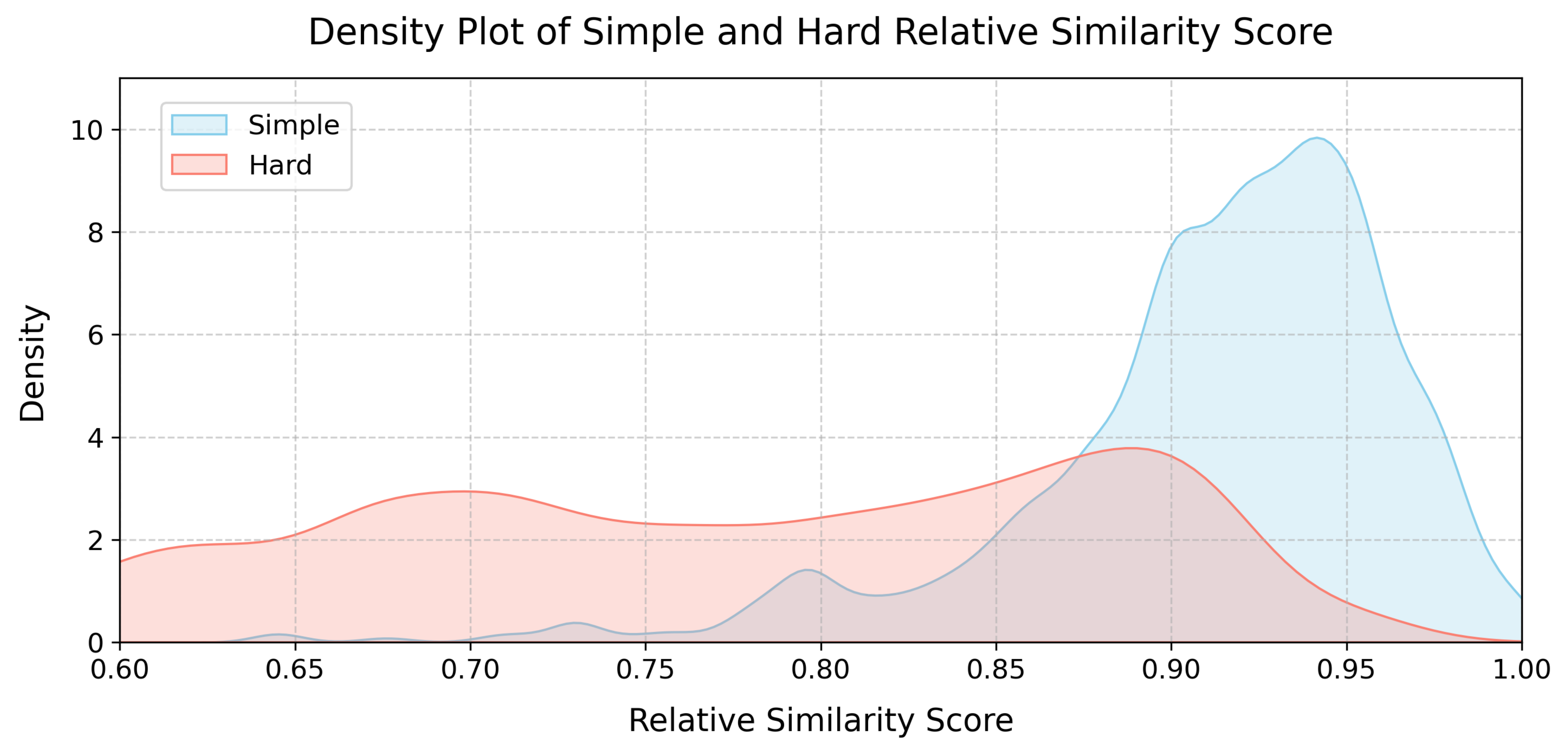}
\caption{Density distribution of Simple (skyblue) and Hard (salmon) images based on Relative Similarity Score, which is the similarity between an image and its class label, normalized by the highest similarity within that class. The vertical axis represents the probability density estimation.}
\label{fig:Similarity_compare}
\end{figure}

\begin{figure*}[tp]
\centering
  \includegraphics[width=\linewidth]{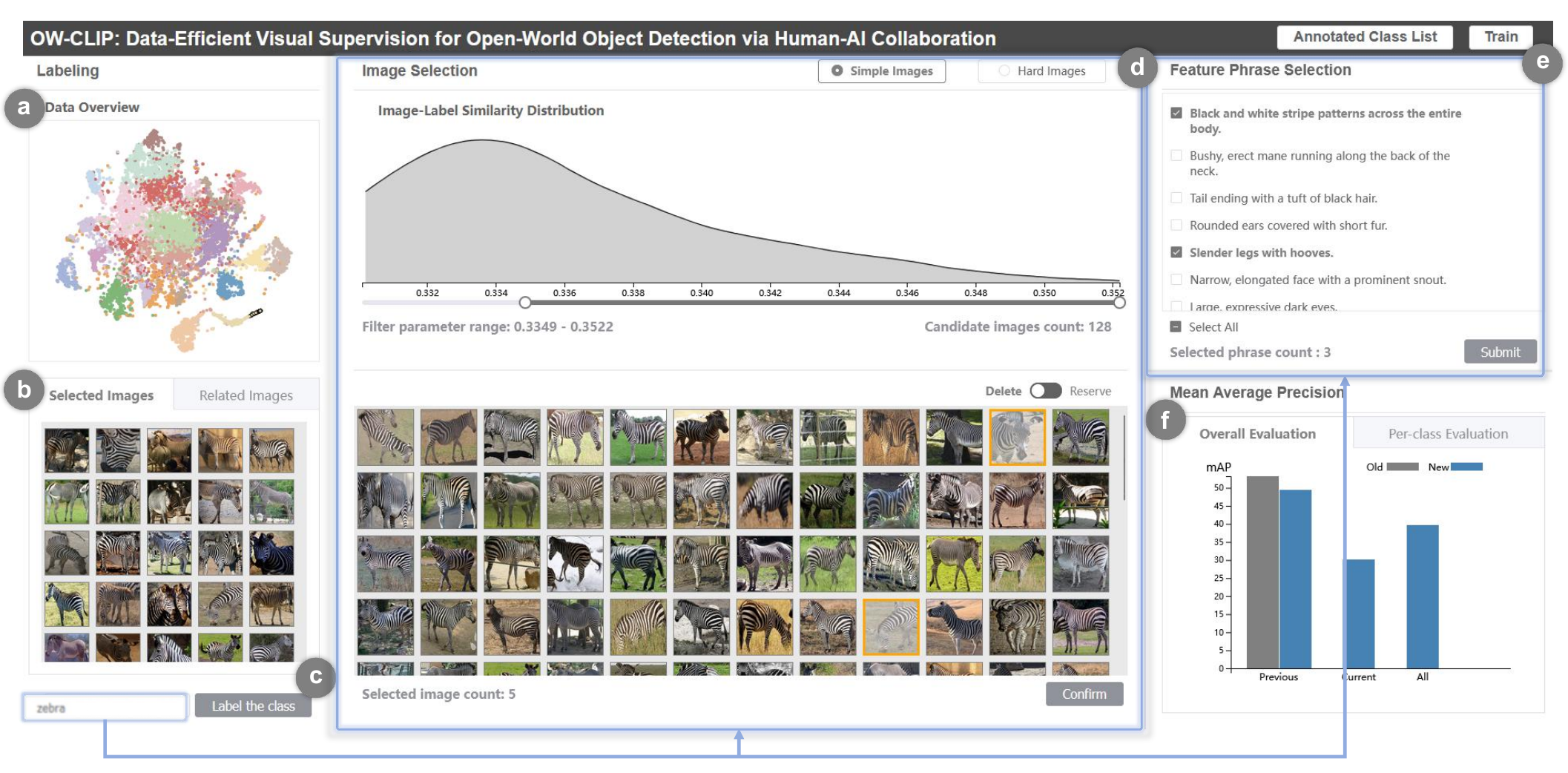}
  \caption{
  	Annotation interface for OW-CLIP. (a) Cluster scatter plot of all unknown detection proposals. User can perform lasso selection on the point clusters, and the images corresponding to the selected points will be displayed in (b). (c) Identify the label name of the potential unknown class. (d) Image Selection View: perform image annotation of Hard image and Simple image. (e)  Feature Phrase Selection: select the appropriate visual feature phrases in GPT-4o’s answers. (f) The evaluation results of the model are displayed, including overall and individual.
  }
  \label{fig:teaser}
\end{figure*}

In conjunction with textual annotation, image data plays a pivotal role in the incremental learning paradigm of the CLIP classifier. In our methodological approach, we systematically differentiate and filter detection proposals for each target class by categorizing all annotated images into two distinct taxonomies: "Simple" or "Hard", defined according to the following criteria:
\begin{itemize}
    \item Simple image: Images that contain the complete object of the specified class with minimal occlusion or background interference.
    \item Hard image: Images that capture only partial object representations or contain significant background elements in addition to the primary object.
\end{itemize}

\textit{Cross-Modal Similarity Guided Image Filtering.}
The effective classification and annotation of images present significant challenges in open-world object detection systems, particularly when dealing with large volumes of unlabeled data.
To address this fundamental challenge, we conducted a rigorous preliminary experiment involving the annotation of approximately 1,200 Simple and Hard images across diverse object classes. The subsequent analysis of their distribution, as illustrated in Figure~\ref{fig:Similarity_compare}, revealed a notable pattern: despite the pre-trained CLIP model occasionally assigning higher classification confidence scores to Hard images than to Simple images, the majority of Simple images consistently achieved superior confidence scores overall. This empirically derived observation serves as the foundational principle for our data filtering methodology. 
Specifically, we leverage similarity threshold ranges, derived from the original CLIP model and the corresponding class labels, to systematically identify distinct \textit{candidate sets} for each image category.
The similarity between an image \( i \) and a class label \( c \) is computed as:
\begin{equation}
    \label{eq:similarity_corrected}
    s(i) = \frac{f^{\text{img}}(i) \cdot f^{\text{txt}}(c)}{\| f^{\text{img}}(i) \| \, \| f^{\text{txt}}(c) \|}
\end{equation}
where \(f^{\text{img}}\) and  \(f^{\text{txt}}\) denote the image encoder and text encoder of original CLIP model, respectively. The filtering parameters can be dynamically adjusted by users based on their specific needs.


\subsection{Human-AI Collaborative Annotation Interface}
As illustrated in Figure~\ref{fig:teaser}, our collaborative annotation system merges several interactive views into a unified interface to optimize human-AI synergy. Building on the data refinement methods in last section, the annotation interface enables target class identification, feature phrase selection and fine-grained image differentiation. Users can complete data annotation within a coherent workflow by seamlessly integrating these components.
\subsubsection{Target Class Identification}
\label{sec:Label-Recommend}

The \textit{Labeling View} serves as the starting point for the annotation process, where users are required to identify potential unknown classes and assign appropriate names to them. To help distinguish between known and unknown classes, users can refer to the Annotated Class List, which displays all previously labeled classes. The criterion for determining whether a class can be annotated is \textit{whether the unknown image set contains enough images of that class for training}. The visualization interface provides two approaches to explore these unknown classes: Cluster Recommendation and Related-Image Recommendation.

\begin{figure}[t]
  \centering 
  \includegraphics[width=0.95\columnwidth]{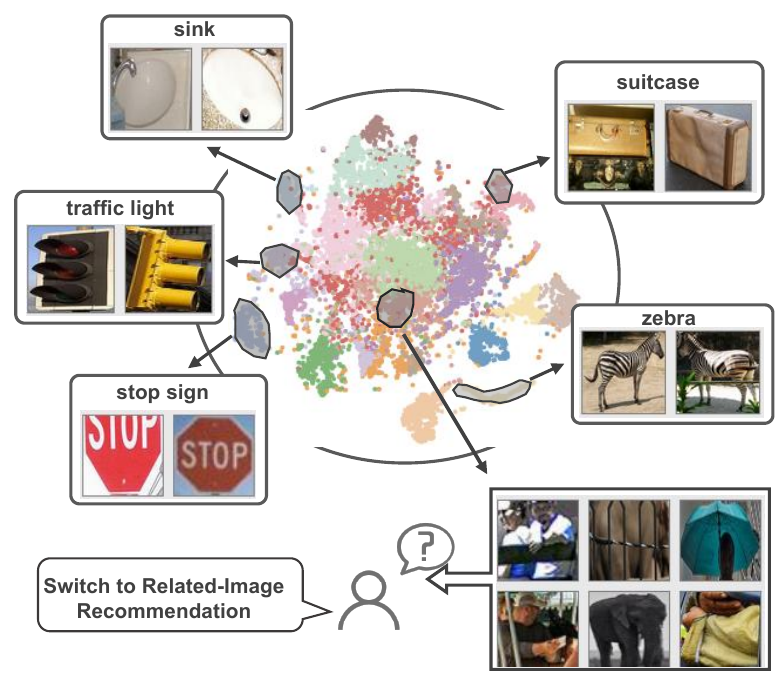}
  \caption{Cluster Recommendation. The lasso tool was used to inspect images in different regions, quickly identifying classes of interest.
  }
  \label{fig:Cluster}
\end{figure}

\textit{Cluster Recommendation.}
The system utilizes image encoder to process and encode all unknown proposals. The encoded features are then clustered using K-means~\cite{hamerly2003learning}, which offers high computational efficiency and is well-suited for high-dimensional embeddings~\cite{caron2018deep}. The optimal number of clusters is determined using standard metrics such as the Sum of Squared Errors (SSE)~\cite{thinsungnoena2015clustering}. 
The cluster result is dimensionally reduced using t-SNE~\cite{van2008visualizing}, which is known to emphasize inter-cluster distances in low-dimensional space. This property enhances the visual separability of semantic groups, making it easier to distinguish between different clusters~\cite{xia2021revisiting}. 
Users can interact with the scatter plot by selecting arbitrary areas, revealing the images corresponding to the selected points, as shown in Figure~\ref{fig:Cluster}. 
This mechanism allows rapid visualization of similar image clusters, helping users identify potential unknown class labels quickly.

\begin{figure}[t]
  \centering 
  \includegraphics[width=0.9\columnwidth]{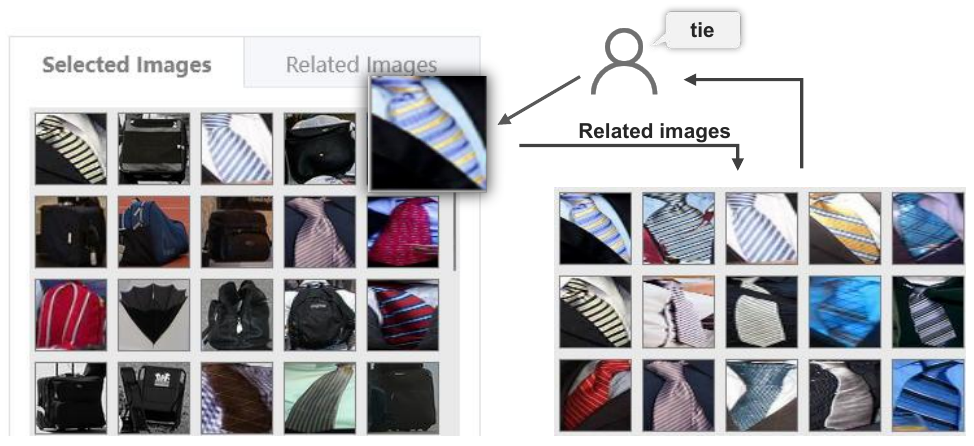}
  \caption{Related-Image Recommendation. Click on a "tie" image that may belong to a potential unknown class,  and the system will display images that are highly similar to the selected image in the Related Images View. 
  }
  \label{fig:crop_SimilarImage}
\end{figure}

\textit{Related-Image Recommendation.}
The effectiveness of the Cluster Recommendation method relies heavily on the quality of the clustering and dimensionality reduction algorithms. Suboptimal results can lead to mixed clusters in the scatter plot (e.g., central region of Figure~\ref{fig:Cluster}). The Related-Image Recommendation module provides an alternative approach for discovering potential unknown classes, as shown in Figure~\ref{fig:crop_SimilarImage}.
Initially, a random subset of images from the unknown cluster is displayed. When the users suspect that a particular object, \( A \) , they can click on the image, and the system computes CLIP-based visual similarity between \( A \)  and all other images by extracting features using the CLIP image encoder. Cosine similarity is used to retrieve the top-100 most similar images, which are then displayed.  Users can quickly assess the number of similar images to determine whether \( A \)  represents a viable unknown class for annotation.

\subsubsection{Dual-Modal Data Annotation}
\label{sec:Superior-Annotation}

After users enter an unknown class label in the designated input box (Figure~\ref{fig:teaser}(c)), the system divides the candidate data into two separate views, specifically designed for the visualization and annotation of textual and image modal data respectively.

The \textit{Feature Phrase Selection View} offers a clear and interactive interface where LLM-generated feature phrases for the target class are listed. Due to the inherent hallucination tendencies of LLMs~\cite{huang2025survey,bai2024hallucination,xu2024hallucination}, not all generated phrases accurately represent the distinctive characteristics of a given class. 
For instance, a LLM might generate the phrase "graceful, slow-paced walking motion" for the class "Giraffe", which fails to capture the distinguishing features that differentiate giraffes from other taxonomic categories. 
To address this, each phrase is accompanied by an interactive checkbox, enabling users to quickly confirm or reject the phrases that best represent the visual attributes of the class. 
By presenting the phrases in a structured, selectable format, the interface reduces the manual effort required to identify relevant feature descriptions.

Building upon image filtering foundation, we developed the \textit{Interactive Image Selection View}, a specialized interface where users can efficiently annotate both Simple and Hard image categories. To facilitate comprehensive understanding of the annotation landscape, the distribution of similarity scores is visualized through a density plot, where the area under each curve represents the frequency of images corresponding to specific score ranges. Users can dynamically adjust the filtering thresholds (\(l_s, h_s, l_h, h_h\)) via interactive sliders, which in turn update the density plot and candidate sets in real time.  
When navigating between image types, users can seamlessly switch to the corresponding view. Notably, the system implements contextually appropriate actions—"Delete" and "Reserve"—for each respective view: 1) For Simple images, where only some may not meet the criteria, selected images are removed from the candidate set, while the rest are kept for further training. 2) For Hard images, which are more challenging to identify, selected images are directly added to the training dataset.



\subsection{Multimodal Supervision with Curated Data}
Each coherent annotation workflow yields Curated training data for a specific class. Once Curated annotated data for multiple unknown classes have been collected, we employ contrastive learning for multimodal prompt tuning, as illustrated in Figure~\ref{fig:overview}.

\textit{Text Modality.} The selected feature phrases are encoded using the CLIP text encoder to derive their embedding representations. Since users often select multiple phrases for single class, the encoded vectors are averaged to initialize the continuous text vector representation, thereby capturing the comprehensive semantic attributes of the target class. This aggregated vector encapsulates the distinctive visual features described in the phrases.

\textit{Image Modality.} A dual-path Crop-Smoothing processing strategy is employed to handle both Simple and Hard image. For Simple images, the hyperparameter $D$ is set to 1.0, indicating full classification confidence. In contrast, Hard images represent incomplete objects and have $D$ values less than 1.0. Hard images are crucial to our system as they help align the distributions of transformed images with the true characteristics of image data. Images produced via Crop-Smoothing often show regularities that differ from the actual data distribution. To better align these distributions, we incorporate the original partial object region images generated by the detection proposal cropping model, referred to as Hard images, as part of our training dataset.

\begin{table*}[th]

\centering
\begin{tabular}{l|cc|cccc}
\toprule

\multirow{3}{*}{\centering Method} & \multicolumn{2}{c|}{Task-1} & \multicolumn{4}{c}{Task-2} \\
\cmidrule(lr){2-7}
 & mAP & \multirow{2}{*}{\centering \#Annotations} & \multicolumn{3}{c}{mAP} & \multirow{2}{*}{\centering  \#Annotations} \\
 & Current Known& & Previously Known & Current Known & Both& \\
\midrule
Faster-RCNN~\cite{faster2015towards} & 56.4 & \multirow{4}{*}{\centering 47223} & 3.7 & 26.7 & 15.2 & \multirow{4}{*}{\centering 113741} \\
Faster-RCNN + Finetuning & 56.4 & & 51.0 & 25.0 & 38.0 & \\
DDETR~\cite{zhu2020deformableAdapter} & 60.3 & & 4.5 & 31.3 & 17.9 & \\
DDETR + Finetuning & 60.3 & & 54.5 & 34.4 & 44.8 & \\
\midrule
ORE~\cite{joseph2021towards} & 56.0 & \multirow{4}{*}{\centering 47223} & 52.7 &26.0 & 39.4 & \multirow{4}{*}{\centering 113741}\\
OST~\cite{yang2021objects} & 56.2 & & 53.4 & 26.4  & 39.9 & \\
OW-DETR~\cite{gupta2022ow} & 59.3 & & 53.6 & 33.5 & 42.9 & \\
RandBox~\cite{wang2023random} & 61.8 & & 54.8 & 35.7 & 45.2 & \\
RandBox-S & 46.5 &4726 & 38.2 & 29.3 & 33.7 &5528 \\
\midrule
Our-Method-S & 51.3 & 1297 & 46.2 & 28.4 & 37.3 & 1033 \\
Our-Method & 53.2& 4427 & 49.6 & 30.3 &40.0 & 4391 \\
\bottomrule
\end{tabular}

\caption{Model Performance Comparison. Previously Known and Current Known refers to the classes the model has learned in the past and the newly acquired classes, respectively. The standard object detectors ( Faster R-CNN and DDETR ) achieve promising mAP for known classes but are inherently not suited for the OWOD setting since they cannot achieve dynamic growth. In the OWOD setting, we compare several advanced models, among which \textbf{RandBox-S} represents the performance of the SOTA model on the same amount of data as our method. \textbf{Our-Method} denotes the performance achieved with the optimal annotation quantity. \textbf{Our-Method-S} denotes the performance achieved with a reduced amount of annotations.}
\label{tab:main_result}
\end{table*}

\section{Experiments}
We performed an experiment to quantitatively evaluate the performance of the developed model incremental traning method. A case study was conducted to indicate the usefulness of visualization system.

\subsection{Quantitative Evaluation on OWOD}
This experiment aimed to evaluate the effectiveness of our OWOD model training utilizing curated annotations.

\textit{Dataset.} 
Following the experimental setting of the conventional OWOD task in the past, we used the classes and images in the Pascal VOC~\cite{everingham2010pascal} and MS-COCO~\cite{lin2014microsoft} datasets. 
We used 40 classes for experiments. In Task-1, the known set $K$ contains the 20 classes in Pascal VOC, and the unknown set $U$ contains the 60 classes unique to MS-COCO. In Task-2, 20 classes in $U$ are added to $K$, which are about Outdoor, Appliance, Acessories, Truck.
It should be noted that we processed and annotated images from the conventional OWOD dataset using our visualization system, the annotated results were used as our training data, which differ from the original dataset's train instances.

\textit{Evaluation Metrics.} 
In order to quantify incremental learning capability of the model in the presence of new labeled classes, we measured the mean Average Precision (mAP)~\cite{everingham2015pascal} at IoU threshold of 0.5. At the same time, we also compared the scale of training data used in the past conventional OWOD task with the scale of data annotated by our visualization system.

\textit{Implementation Details.}
We used Faster R-CNN model~\cite{faster2015towards} as our detection proposal generator.
The model consists of two stages: a region detection network (RPN) and a region of interest (RoI) classification head. At the same time, we used GPT-4o~\cite{GPT-4o} as the interactive LLM. During training, all experiments we provided were based on the pre-trained CLIP ViT-B/16 model~\cite{radford2021learning}. When we applied multimodel prompt tuning on CLIP, we set vision prompt lengths to 10 , and the text encoding results of all feature phrases were used as the initial value of the context prompt. All models were trained for 20 epochs with a batch-size of 32 and a learning rate of 0.02 on NVIDIA-3090 GPUs.

\textit{Result.} Table~\ref{tab:main_result} presents the evaluation of OW-CLIP against advanced methods using strong supervision in the standard OWOD dataset, observing the following findings:
1) Although using only 9.3\% and 3.8\% of the training data in Task-1 and Task-2, respectively, OW-CLIP achieved more than 85\% state-of-the-art (SOTA) performance. In Task-2, our method even reached 89\% of SOTA performance, outperforming most existing models. 2) To ensure a more fair comparison, we trained RandBox-S on a randomly sampled subset of the original RandBox dataset, using a similar number of training images and a balanced class distribution as in OW-CLIP. Our method
significantly outperformed RandBox-S, validating the effectiveness of
our human-supervised annotations and training approach. 3) After adding 20 classes in Task2, the forgetting rate of the previous classes was only 8\%, while the forgetting rate of the SOTA model reached 12\%, which proves that our method can effectively alleviate the forgetting problem after updating the model.  4) Through analysis, we found that optimal performance was achieved with around 220 images and four feature phrases per class. This required about 4 minutes for simple classes and 6–9 minutes for complex ones, with limited gains beyond this level. 
We also tried fewer annotations (60 images per class, 2-3 minutes annotation time), showed slightly reduced performance (denoted as Our-Method-S),  offering users a practical trade-off between annotation effort and model effectiveness.

We also conducted a comparison with OW-Adapter~\cite{jamonnak2023ow}, an advanced OWOD visualization system.  We invited two experts who had no prior involvement in the development of our system. After providing them with a brief introduction to the visualization interface and annotation guidelines, the experts annotated the 10 unknown classes from the OW-Adapter experimental setup. The amount of annotated data was approximately the same as that of OW-Adapter.
As shown in Table~\ref{tab:case-study}, OW-CLIP outperformed OW-Adapter by 8.9\% overall on unknown classes, with particularly significant improvements for "giraffe" (22.3\%) and "zebra" (20.4\%), further validating our approach.

\begin{table}[tp]
\begin{tabular}{l|cc|cc}
\toprule
\multirow{2}{*}{\centering Method}         & \multicolumn{2}{c|}{Class(\#Annotations)}      & \multirow{2}{*}{Unknown}  & \multirow{2}{*}{\#Annotations}                                                    \\
 & giraffe  & zebra \\
\midrule
\textbf{Our-Method} & \textbf{76.1(81)}    &  \textbf{72.8(77)}  &\textbf{26.63}   &  \textbf{463}                                                   \\
OW-Adapter             & 53.8(110)     & 52.4(75) & 18.75   & 416                  \\
\bottomrule
\end{tabular}
\caption{The experimental results compared with OW-Adapter. Here, "Unknown" refers to the ten new classes extracted from the MS-COCO dataset, following the experimental setup of OW-Adapter.}
\label{tab:case-study}
\end{table}
\subsection{Case study}
To evaluate the efficiency of OW-CLIP in data annotation and the quality of the resulting annotations, we carry out a case study on the subset of Pascal VOC and MS-COCO. The subset is the same as that we have used in the quantitative evaluation.

\textbf{Annotation Process} 

We assumed that the current object detection model has been pre-trained on Task-1 classes. After processing the input data through the model, detection proposals from unrecognized classes (potential unknowns) are filtered out. 
The user first needed to identifying potential unknown classes. Leveraging the lasso tool on the scatter plot, he selected clusters individually, prioritizing those that are visually distinct and exhibit tightly grouped intra-cluster points. The selected images were displayed in the Selected Images View for a brief review. The user assigned the corresponding label (e.g., "zebra," "giraffe") to the classes which were prominently represented in this view. 
For regions in the scatter plot where clusters were less distinct and mixed with other classes, the user turned to the Related-Image Recommendation. He clicked on the image that might be a potential class, the system retrieved visually similar images and displayed them in the related image view. This method was used to identify classes such as "tie", "oven", "microwave" etc.
Upon class label confirmation (e.g., "zebra"), the system presents candidate Simple and Hard images in the Image Selection View, while simultaneously displaying 10 GPT-4o-generated visual feature phrases for the class.
For Simple image annotation, the user adjusted parameters $ls$ and $hs$ to 0.3349 and 0.3522, yielding 128 candidate images. Using the "Delete" mode, the user removed inappropriate images, retaining 123 images as training data. This process was then repeated for Hard images with appropriate threshold adjustments.
The user then selected 3 of the 10 displayed visual feature phrases as the definitive textual representation, completing the annotation for the "zebra" class.
This procedure was applied to all 20 classes in Task-2. 



\begin{table}[t]
\begin{tabular}{l|c|ccc}
\toprule
\multirow{3}{*}{\centering Method}              & Task1         & \multicolumn{3}{c}{Task 2}                                                        \\
\cmidrule(lr){2-5}
 & mAP           & \multicolumn{3}{c}{mAP}                                                           \\
                    & C-Known & P-Known & C-Known & Both \\
\midrule
\textbf{Our-Method} & \textbf{53.2 }         & \textbf{49.6}                                                       & \textbf{30.3}          & \textbf{40.0} \\
\midrule
\textit{wo-PhraseSelection}              & 51.9         & 48.1                                                     & 28.5          & 38.3 \\
\textit{wo-LLM}              & 51.1          & 48.3                                                       & 26.1          & 37.2 \\
\midrule

\textit{wo-Differentiation}    & 49.4       & 45.9                                                & 27.6         & 36.8\\
\textit{wo-CS}    & 43.4          & 42.3                                                       & 24.4          & 33.3\\

\bottomrule
\end{tabular}
\caption{The table presents the results of progressive data degradation experiments, where C and P denote the current and previous classes, respectively. \textit{wo-PhraseSelection} indicates the use of all phrases generated by the LLM without manual filtering, while \textit{wo-LLM} refers to using only class label names. \textit{wo-Differentiation} skips the manual classification of images into Simple and Hard, and \textit{wo-CS} eliminates Crop-Smoothing and uses crowdsourced images. }
\label{tab:Ablation}
\end{table}

\textbf{Annotation Quality}

To evaluate the impact of our visualization system on annotation quality, we conducted progressive data degradation experiments, with results shown in Table~\ref{tab:Ablation}. \textit{Textual Annotation Degradation:} we first removed manual phrase selection, relying solely on LLM-driven phrase generation. This change led to modest decreases in mAP—by 1.4\% for Task-1 and 1.7\% for Task-2—demonstrating the added value of manual filtering. Subsequently, we eliminated all feature phrases and replaced them with the default context "A photo of [class]," which resulted in further mAP reductions of 2.1\% and 2.8\% for Task-1 and Task-2, respectively. \textit{Visual Annotation Degradation:} we initially removed manual image differentiation by using only Simple image , which resulted in mAP declines of 3.8\% for Task-1 and 3.2\% for Task-2.  Next, we replaced all manually annotated images with those from crowdsourced datasets, and Crop-Smoothing was eliminated in favor of traditional one-hot encoding; under these conditions, mAP dropped significantly by 9.8\% and 6.7\% for Task-1 and Task-2, respectively.
These results underscore the importance of both manual supervision and advanced data processing techniques in enhancing the quality of annotated data for model training.

To further understand the trade-off between annotation quality and user effort, we analyze the annotation time and performance impact. Although the performance gain from LLM-based phrase filtering is relatively modest (~1.8\%), the interaction cost is extremely low—users only need to make simple binary decisions, typically within 20–30 seconds per class. In contrast, the Image Differentiation module yields a larger improvement (3.2\%) with slightly more effort. Importantly, both tasks require only basic common sense, making the system accessible to non-expert users and practical for large-scale annotation.

\begin{table}[t]
\renewcommand{\arraystretch}{1.2}
\centering
\begin{tabular}{l|c|c}
\toprule
\textbf{Metric} & \textbf{Description}  & \textbf{Mean ± Std} \\
\midrule
Annotation Time & Avg. time per class  & 3.6 ± 1.1 \\
Perceived Ease of Use & Avg. usability score & 5.5 ± 0.6 \\
Cognitive Load & Avg. workload score  & 2.6 ± 0.8  \\
\bottomrule
\end{tabular}
\caption{User study results for the system. Annotation Time is in minutes. Perceived Ease of Use and Cognitive Load are rated on a 7-point scale, where 7 indicates highest ease and highest load, respectively.}
\label{tab:user-study}
\end{table}

\textbf{Comprehensive User Study}

To assess the usability and efficiency of our system, we conducted a comprehensive user study involving 20 participants with no prior experience using the tool. After a brief introduction to the system’s functionality and interaction workflow, and following the evaluation setting adopted in OW-Adapter, each participant was asked to annotate 10 object classes using our system. 

After completing the annotation tasks, the system recorded each participant’s average annotation time, and we administered a post-task questionnaire assessing two key aspects: perceived ease of use and cognitive load.
\textit{Perceived ease of use} was evaluated adapted from the Technology Acceptance Model (TAM)~\cite{davis1989perceived} and customized to fit our visualization system. The assessment focused on five key aspects: ease of learning, logical interaction design, timeliness of feedback, interaction flexibility, and smoothness of the annotation process.
\textit{Cognitive load} was assessed based on the NASA Task Load Index (NASA-TLX)~\cite{hart1988development} framework, with modifications to reflect our annotation workflow. We selected three task-relevant dimensions to measure  workload: target class identification, image differentiation, and feature phrase selection.

The results are summarized in Table~\ref{tab:user-study}. The average perceived ease of use score was 5.5 out of 7 (±0.6), suggesting that users found the system intuitive and easy to operate, even without prior training. At the same time, the average cognitive load was relatively low at 2.6 (±0.8), indicating that key annotation tasks required limited mental effort. The average annotation time per class was 3.6 minutes (±1.1), which is already efficient given that all participants were first-time users. We expect that with minimal training or repeated use, this time could be further reduced. These findings highlight the system’s accessibility and low cognitive demand, making it well-suited for practical use at scale, especially by non-expert users.


\section{EXPERT FEEDBACK AND DISCUSSION}

\subsection{Expert Feedback}

We invited three experts from the field of computer science (E6, E7, E8) to assess our OW-CLIP system. These experts are masters or Ph.D. researchers specializing in distinct areas, and have more than two years of research experience. 
It should be noted that these experts are not collaborators in the development of our system.
We first introduced them to the purpose and function of the system. Then, the each feature of the system were demonstrated to them step by step.  After that, we conducted a interview with them and summarized feedbacks as follows:

\textit{Reducing Effort and Interaction Simplicity.}
The experts particularly praised the low effort required. E6 remarked, "Transforming the LLM's text-stream responses into checkbox lists is a wonderful operation. I don’t have to write anything—I just click. It is accurate and easy to scan through," E8 also appreciated the Image-Label Similarity Distribution visualization, stating, "If I know the amount of annotation in advance, I can determine the approximate parameter values accordingly."

\textit{Flexible Class Search.}
Our scatter plot visualization presents clustered image features in 2D, enabling users to flexibly define classes by selecting free-form regions. This greatly simplifies the discovery of new classes without requiring prior knowledge. Expert E6 emphasized its usability: "The layout gives me a clear sense of how the data is grouped semantically. It's easy to spot which clusters are meaningful." 


\textit{Model Performance.}
Our model’s performance left a strong impression on E7, he commented, "Achieving results comparable to advanced models with less than 5\% of the data is remarkable. Although the training data was carefully curated, I believe it was well worth the effort. With the help of this system, annotating data has become a much easier task. " It opens up new possibilities for deploying high-performing models in environments where data collection is limited. 


\subsection{Limitation and Future Work}

Despite the positive feedback, limitations of OW-CLIP were identified.


\textit{Class Feature Learning Across Episodes.} 
Our method employs contrastive learning to train multiple classes, where each training episode only captures similarities and differences between samples within that episode. However, the model cannot distinguish between classes from different episodes,thereby requiring careful and strategic class assignment across episodes. At the same time, this training strategy also implies that if annotation errors occur for a specific class, it necessitates re-training all classes within the same episode. 

\begin{figure}[t]
  \centering 
  \includegraphics[width=0.93\columnwidth]{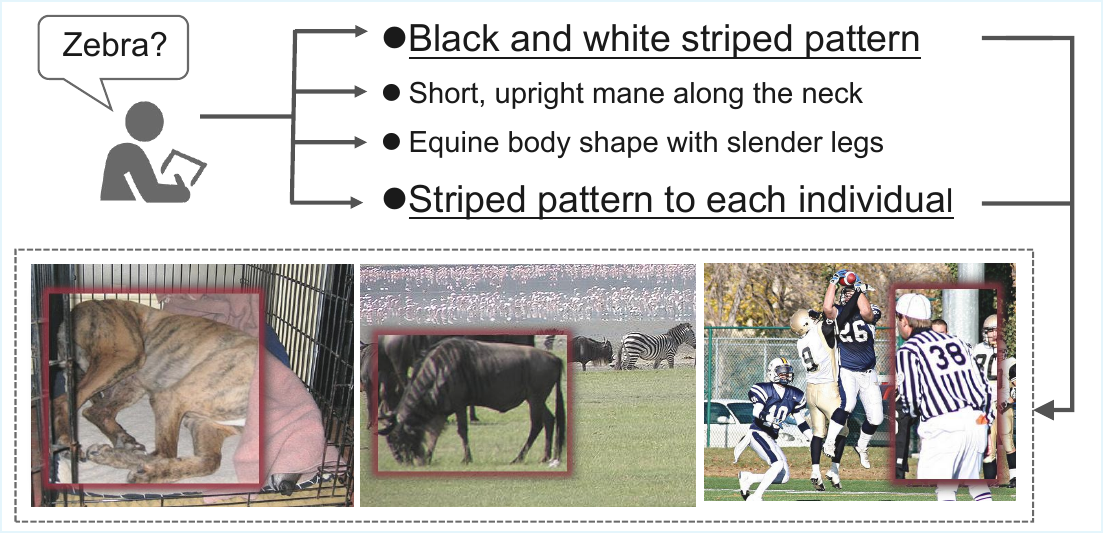}
  \caption{An example of misguidance caused by feature phrases.
  }
  \label{fig:badcase}
\end{figure}
\textit{Misguidance from Feature Phrases.} Our method relies on visual feature phrases to guide model learning, which introduces potential risks when the selected phrases emphasize misleading attributes. As illustrated in Figure~\ref{fig:badcase},   in the case of "zebra" annotation, phrases "black and white striped pattern" and "striped pattern unique to each individual" both highlight surface texture as a dominant cue. This redundancy can lead the model to overfit to superficial features ,  causing incorrect predictions on visually similar but semantically distinct classes.

\textit{Multi-turn Conversations with LLM.}
E6 suggested that "Large language models excel at multi-turn conversations, allowing them to correct responses based on historical exchanges. If I find that none of the generated feature phrases are satisfactory, I would need the LLM to refine its answers." He highlights that this situation is particularly likely to occur with models that are performance-constrained.

\section{CONCLUSION}

In this work, we present OW-CLIP, a visual analytics system that addresses three fundamental limitations in open-world object detection: data quality scarcity, partial feature overfitting, and model inflexibility. Through human-AI collaboration, OW-CLIP introduces three key innovations: 1) a plug-and-play prompt tuning strategy tailored for OWOD settings; 
2) curated data, including Visual Feature Phrases and Fine-Grained Differentiated Images; 3) the Crop-Smoothing technique, mitigating partial feature overfitting by dynamically adjusting predictions based on object completeness;
Quantitative evaluation experiment demonstrates that OW-CLIP achieves competitive
performance at 89\% of state-of-the-art benchmark while requiring only 3.8\% self-generated data. When trained with equivalent data
volumes, our framework outperforms current SOTA methods. A case study confirm the effectiveness of the developed method and the improved annotation quality of our system. User study further confirms the usability of our system.



\acknowledgments{%
This paper is supported by National Natural Science Foundation of China (NO. U23A20313, 62372471) and The Science Foundation for Distinguished Young Scholars of Hunan Province (NO. 2023JJ10080). This work was carried out in part using computing resources at the High-Performance Computing Center of Central South University.	%
}

\bibliographystyle{abbrv-doi-hyperref}

\bibliography{template}

\appendix 

\end{document}